# Interpretable Machine Learning based on Functional ANOVA Framework: Algorithms and Comparisons[1]

Linwei Hu[2], Vijayan N. Nair, Agus Sudjianto, Aijun Zhang, and Jie Chen
Corporate Model Risk, Wells Fargo, USA

**Abstract**

In the early days of machine learning (ML), the emphasis was on developing complex algorithms to achieve best predictive performance. To understand and explain the model results, one had to rely on post hoc explainability techniques, which are known to have limitations. Recently, with the recognition that interpretability is just as important, researchers are compromising on small increases in predictive performance to develop algorithms that are inherently interpretable. While doing so, the ML community has rediscovered the use of low-order functional ANOVA (fANOVA) models that have been known in the statistical literature for some time. This paper starts with a description of challenges with post hoc explainability and reviews the fANOVA framework with a focus on main effects and second-order interactions. This is followed by an overview of two recently developed techniques: Explainable Boosting Machines or EBM (Lou et al., 2013) and GAMI-Net (Yang et al., 2021b). The paper proposes a new algorithm, called GAMI-Lin-T, that also uses trees like EBM, but it does linear fits instead of piecewise constants within the partitions. There are many other differences, including the development of a new interaction filtering algorithm. Finally, the paper uses simulated and real datasets to compare selected ML algorithms. The results show that GAMI-Lin-T and GAMI-Net have comparable performances, and both are generally better than EBM.

## 1 Introduction

Interpreting the results from machine learning (ML) algorithms has been the subject of considerable interest. Early approaches relied on post hoc techniques, i.e., techniques to explain the results of a model after it has been fit. They include permutation-based variable importance (Breiman 1996), partial dependence plots (Friedman 2001), and H-statistics (Friedman and Popescu 2008). Many of these are low-dimensional summaries of high-dimensional models with complex structure, and hence can be inadequate or even misleading. A second approach for model explanation is the use of surrogate models (or distillation techniques) that fit simpler models to extract key information and use that to explain the original complex models. Examples include: i) LIME (Ribeiro et al., 2016) based on local linear models; and ii) locally additive trees for global explanation (Hu et al., 2020). However, these are only approximations and may not be "true" to the original model.

A more recent direction is the use of ML algorithms to fit inherently interpretable models. Sudjianto and Zhang (2021) described several criteria that interpretable models should satisfy. Among these, we focus on parsimony, which implies having as few active effects and complicated relationships as possible subject to achieving reasonably good predictive performance. Recent papers accomplish parsimony by fitting functional models with just main effects and second order interactions. The rationale for this approach is that, in many areas with tabular data, nonparametric models with lower-order interactions are sufficient in capturing most of the structure.

[1] The views expressed in this paper are those of the authors and do not necessarily reflect those of Wells Fargo.
[2] Corresponding author's email: Linwei.Hu@wellsfargo.com

Of course, this (renewed) emphasis on parsimony will not come as a surprise to the statistics community. The notion of fitting low-order functional ANOVA (fANOVA) models has been around in the statistical literature. Early papers include the pioneering work done by Stone, Wahba, and her students (Stone 1994, Gu 2013). What is new is the use of ML architecture (boosted trees and neural networks) to develop scalable algorithms.

The additive index model (AIM)

$$g(\boldsymbol{x}) = g_1(\boldsymbol{\beta}_1^T\boldsymbol{x}) + g_2(\boldsymbol{\beta}_2^T\boldsymbol{x}) + \ldots + g_K(\boldsymbol{\beta}_K^T\boldsymbol{x}),$$

is one way to generalize GAM to capture certain types of interactions. It was first proposed by Friedman and Stuetzle (1981) as an exploratory tool in the early days of nonparametric regression and was called projection pursuit. Vaughan et al. (2018) showed how one can use restricted neural networks to fit AIMs using gradient-based training. They referred to their method as explainable neural networks (xNNs). Yang et al. (2021a) further enhanced the interpretability of xNNs through architecture constraints.

This paper deals with another class of models based on low-order functional ANOVA (fANOVA). In particular, we focus on main effects (GAMs) and second-order interactions:

$$g(\boldsymbol{x}) = \sum_j g_j(x_j) + \sum_{j \neq k} g_{jk}(x_j, x_k). \qquad (1)$$

This class was referred to as GA2M in Lou et al. (2013), but we use the term GAMI (GAM + Interactions). Explainable boosting machine or EBM (Lou et al., 2013) uses gradient boosting with piecewise constant trees to fit GAMI models. GAMI-Net, developed by Yang et al. (2021b), uses restricted neural-network structures and associated optimization techniques. In this paper, we introduce a new method that we refer to as GAMI-Lin-T. Like EBM, it is a tree-based method but uses linear fits instead of piecewise constant fits within nodes. A new interaction filtering approach is also introduced in the paper, and it is shown to perform better than the FAST algorithm in Lou et al. (2013).

Before moving to the main points of the paper, it is important to mention the advantages as well as limitations of the fANOVA framework.

a) The major reason for relying on fANOVA decomposition of a model into main effects and interactions is their familiarity with users. Most practitioners have learned about these concepts in the context of linear models, and hence using their functional versions for interpretation comes naturally. But, of course, this is just one way of representing the underlying model, and there are many other alternatives.
b) More importantly, the fANOVA decomposition of an underlying model into main effects and decompositions would not necessarily recover the original model. For example, consider the underlying function $f(x) = |x_1| * x_2$. The fANOVA decomposition will try to represent it as $g(x) = g_1(x_1) + g_2(x_2) + g_{12}(x_1, x_2)$, as a sum of orthogonal main effects and interaction functions. It will not necessarily recover the original function. The fitted main effects are projects of the original function into main effects and the remaining will be recovered as second-order interaction effect.

c) Finally, it is worth reiterating that when there is strong correlation, the underlying model is not identifiable. Consider, for example the model $f(x) = x_1 + x_2 + x_1x_2$, and suppose the two variables highly correlated. The interaction will manifest itself, at least partially, as a quadratic term. If you fit the model in stages, first the main effects first and then interactions, the main effects for $x_1$ and $x_2$ will carve out part of the interaction term and will be partly quadratic. This will lead to a dilution of the estimated second-order interaction terms. We will see examples of this later.

The rest of the paper is organized as follows. Sections 2 and 3 provide overviews of EBM and GAMI-Net respectively. The new algorithm, GAMI-Lin-T, is described in Section 4. Section 5 and 6 compare the three GAMI algorithms with XGBoost on simulated data and a dataset on home mortgages. The paper concludes with some summary comments.

## 2 Explainable Boosting Machine (EBM)

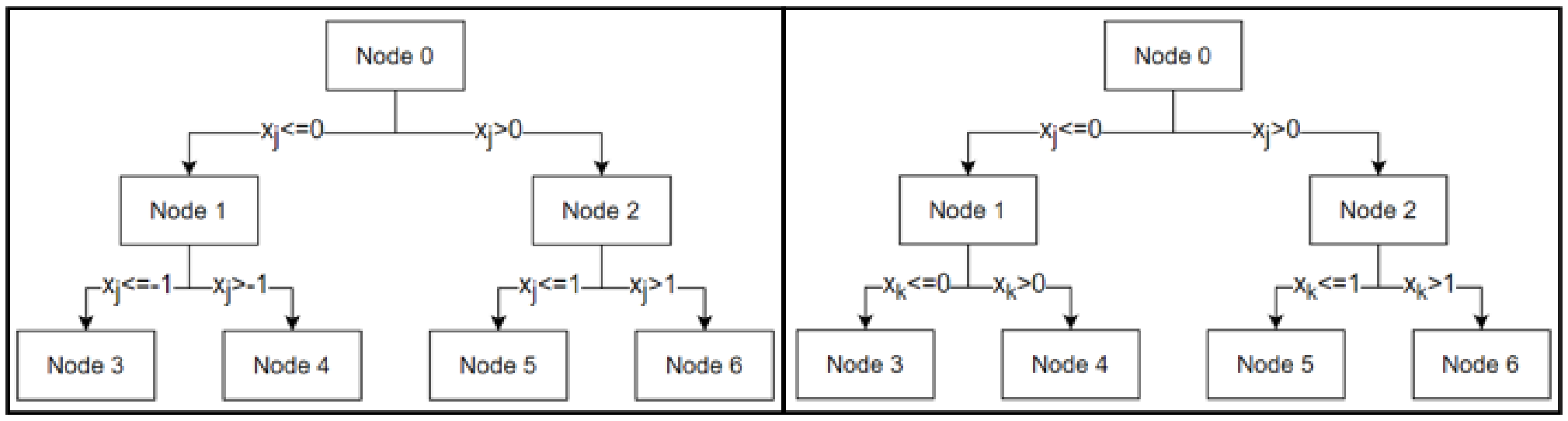

Figure 2-1. Main-effect tree $T(x_j)$ (left) and interaction-effect tree $T(x_j, x_k)$ (right)

This section provides a brief overview of EBM, and readers are referred to Lou et al. (2013) for more details. EBM is a two-stage algorithm where the main effects and two-way interactions in Eq (1) are fitted in stages. In Stage 1, the main effect of each variable is modeled as a function using shallow piecewise-constant trees which split only on that single variable. Stage 2 models the interaction of each pair as a function using shallow trees which split only on that same pair of variables. Figure 2-1 provides a visual representation:

a) The left panel is the tree for main effect, $T(x_j)$. It uses the same variable $x_j$ for all tree splits and fits a constant in each node. Repeated splits on the same variable will model inherent nonlinearities. Boosting allows it to further capture complex nonlinear patterns.

b) The right panel fits an interaction tree, $T(x_j, x_k)$, on the residuals from main effects. The important $(x_j, x_k)$ pairs are identified in a separate step (discussed below). The algorithm first uses variable $x_j$ to split the root node, and then selects the other variable $x_k$ to split the two child nodes, allowing us to model this pairwise interaction.

Within the main effect (or interaction) stage, the algorithm cycles through all variables (or pairs of variables) in a round-robin manner and iterates for several iterations until convergence. Like gradient boosting, each tree is fitted based on residuals from previous set of trees. Further, the interaction trees are fitted based on residuals from the fitted main effects. The algorithm is illustrated in Figure 2-2.

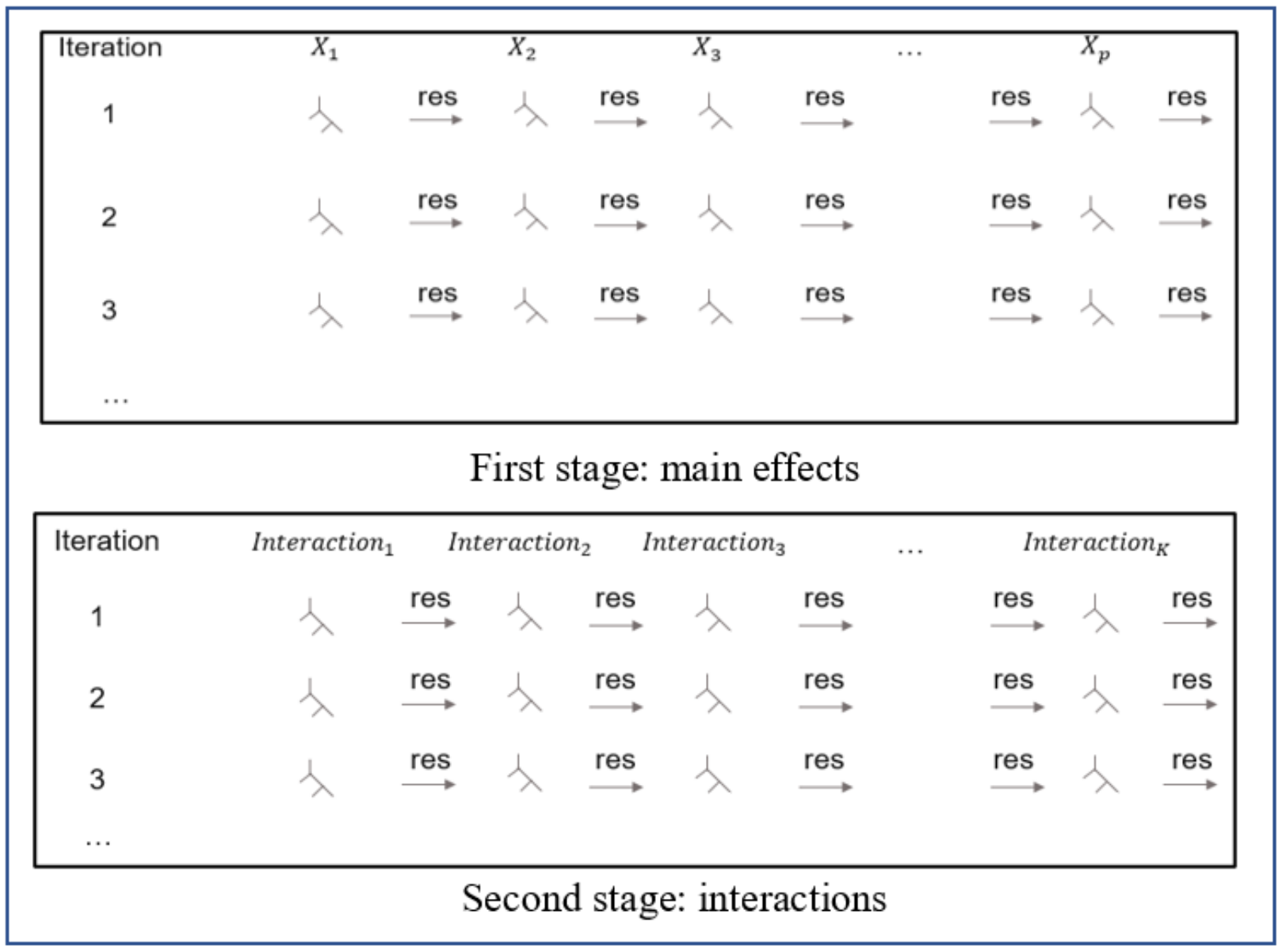

Figure 2-2. First and second stage modeling in EBM

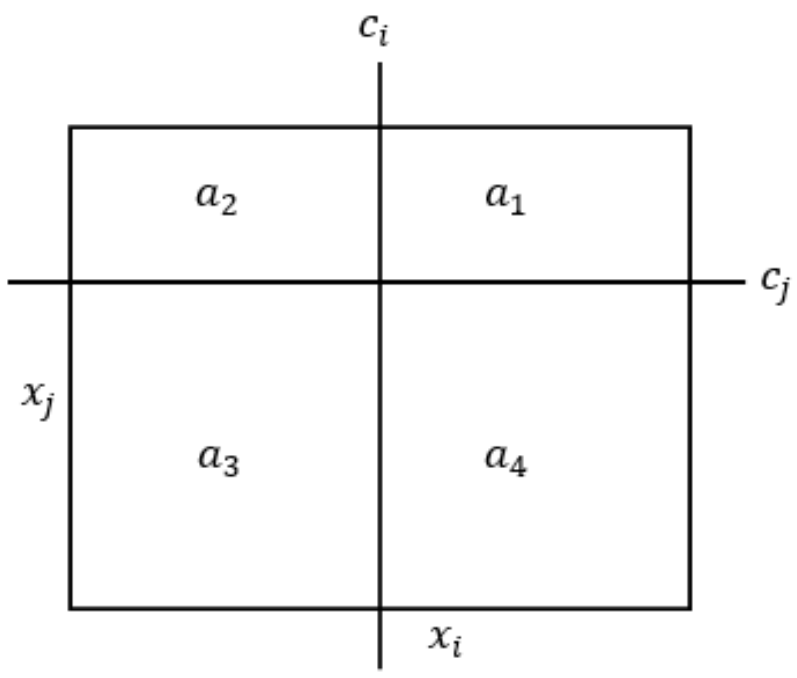

Figure 2-3. Illustration for FAST interaction filtering algorithm

Modeling all possible variable pairs is computationally expensive and also prone to overfitting. So, a filtering method called FAST (Lou et al., 2013) is used to select the top $K$ interactions. Only those K interactions are modeled in the second stage. FAST fits a simple interaction model to the residuals (after removing the fitted main effects) for each pair of variables and ranks all pairs by the reduction in an appropriate metric for model error. The interaction model is a simple approximation which divides the two-dimensional input space into four quadrants and fits a constant in each quadrant to estimate the functional interaction. This process is illustrated in Figure 2-3, where the cut points $c_i$ and $c_j$ are each selected from a set of grid points and the best cut points are chosen to optimize the fit. Lou et al. (2013) justified this approximation by arguing that fully building the interaction structure for each pair "is a very expensive operation".

Once the EBM model is fit, all the main-effect trees for each variable and the interaction-effect trees for each top interaction pair are combined to produce the main-effects and interaction effects in Eq (1). The effect of each term is visualized using a one- or two-dimensional plot, making the model directly interpretable. In addition, the standard deviation of each term is calculated to measure and rank the importance of each term.

# 3 GAMI-Neural-Network (GAMI-Net)

This section provides a brief overview of GAMI-Net, and readers are referred to Yang et al. (2021b) for more details. GAMI-Net is also a multi-stage algorithm that fits the main effects and second-order interactions in Eq (1). Unlike EBM, GAMI-Net uses specialized networks to fit the main and interaction effects.

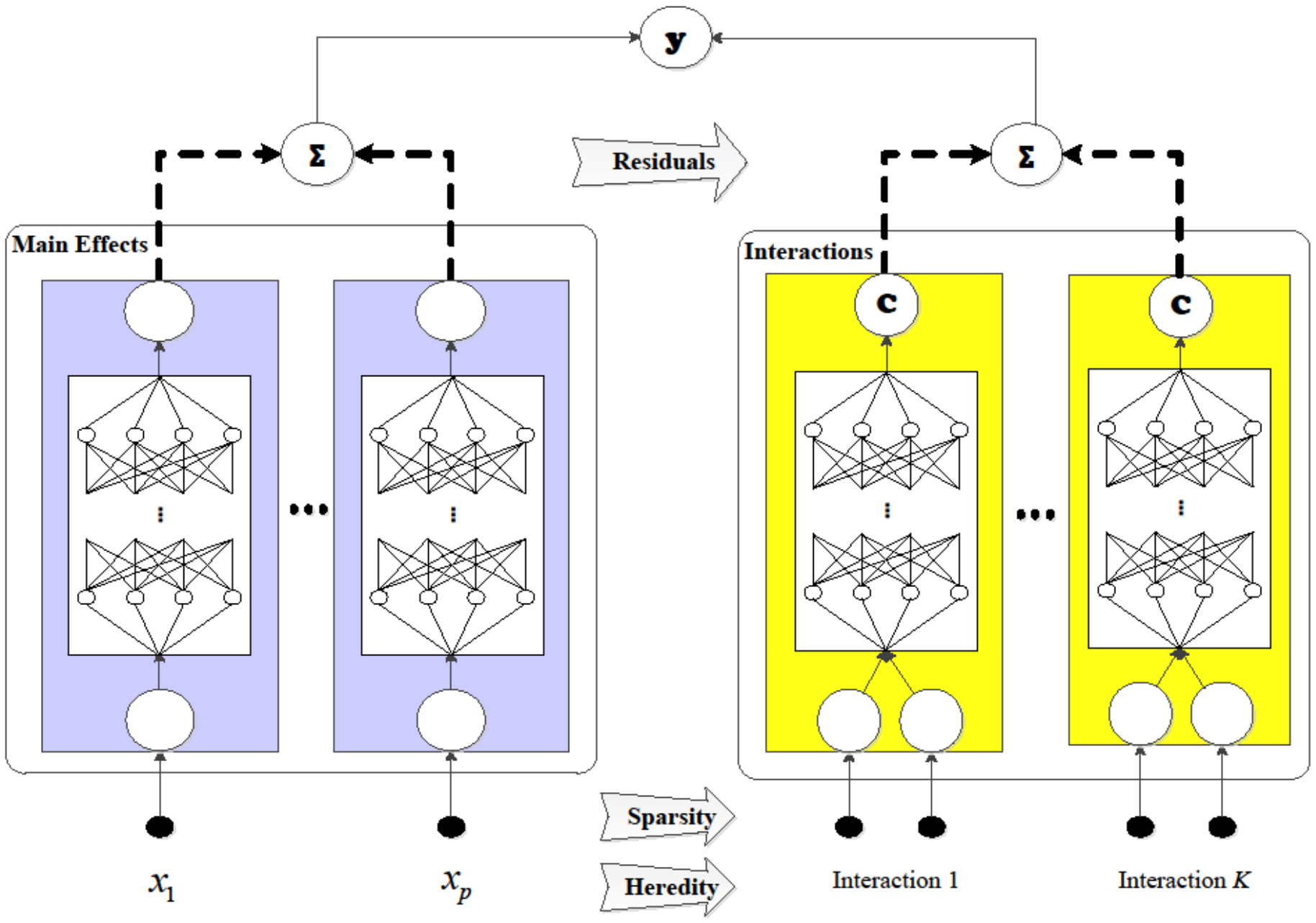

Figure 3-1. The GAMI-Net architecture (Yang et al., 2021b): each main effect is modeled by a subnetwork with univariate input, and each second-order interaction is modeled by a subnetwork with a pair of inputs.

Figure 3-1 shows the network architecture where restricted neural network architecture is used to model the main effects and interactions separately. The original implementation of GAMI-Net by Yang et al. (2021b) incorporated the following interpretability constraints:

a) **Sparsity**: select only the most important main effects and pairwise interactions.
b) **Heredity**: select a pairwise interaction only if at least one (or both) of its parent main effects is selected.
c) **Marginal Clarity**: enforce the pairwise interactions to be nearly orthogonal to the main effects, by imposing the penalty

$$\Omega(h_j, f_{jk}) = \left|\frac{1}{n}\sum h_j(x_j) f_{jk}(x_j, x_k)\right|.$$

In the associated PyTorch package for GAMI-Net, Yang et al. (2021c) added a monotonicity constraint:

d) **Monotonicity:** if monotonicity is required for certain features, make the relationships to be (nearly) monotonic increasing or decreasing, by imposing the penalty

$$\Omega(x_j) = \max\left\{-\frac{\partial g}{\partial x_j},\ 0\right\} \text{(if increasing) or } \max\left\{\frac{\partial g}{\partial x_j},\ 0\right\} \text{(if decreasing).}$$

Note that sparsity and heredity are based on hard constraints, while marginal clarity and monotonicity constraints are soft. These constraints can be naturally added to the objective function in neural network training. GAMI-Net uses a three-stage adaptive training algorithm:

1. Train the main effect subnetworks and prune the trivial ones by validation performance.
2. Train pairwise interactions on residuals, by
   a) Selecting candidate interactions by heredity constraint;
   b) Evaluating their scores (using the same FAST algorithm as in EBM) and select top K interactions;
   c) Training the selected two-way interaction sub-networks; and
   d) Pruning unimportant interactions.
3. Retrain main effects and interactions simultaneously for fine-tuning network parameter.

As with EBM, the fitted effects can be visualized using one- and two-dimensional plots. Yang et al. (2021b) also developed metrics to quantify the importance of each effect. These can be used with all the GAMI models.

# 4 GAMI-Linear-Tree (GAMI-Lin-T)

## 4.1 Methodology

GAMI-Lin-T is also multistage algorithm. As with EBM, it is based on boosted trees and uses separate specialized trees to estimate main-effects and interactions, but there are several key differences. It uses model-based tree which fits linear models within each node instead of the usual piecewise constant tree. See Section 4.3 for other differences. Model-based trees have been used in the literature (M5 (Quinlan 1992), LOTUS (Chan and Loh 2004) and MOB (Zeileis et al., 2008)). As Quinlan (1992) notes, model-based trees often lead to more parsimonious results than piecewise-constant trees. In addition, the fitted response surface is less jumpy. GAMI-Net with ReLU activation function also yield piecewise-linear models within partitions, but the partitions are based on oblique cuts of the predictors (see Sudjianto et al., 2020).

### 4.1.1 Algorithms for Fitting Main Effects and Interactions

Two separate model-based trees are used in GAMI-Lin-Tree to capture main effects and interactions (Figure 4-1):

a) The tree on left panel $T(x_j)$ is for main effects. Unlike EBM, the model $f_v(x_j)$ inside node $v$ is chosen as a simple linear model, with L2 penalty to control overfitting. Repeated splits on the same variable will model inherent nonlinearities.
b) The tree on the right panel $T(x_j, x_k)$ is used to capture interaction effects between these two variables. The model $f_v(x_j)$ inside node $v$ is a linear B-spline model. Modeling the effect of $x_j$ conditional on $x_k$ allows us to capture the interaction among the two variables. We apply L2 penalty to each model to control for overfitting.

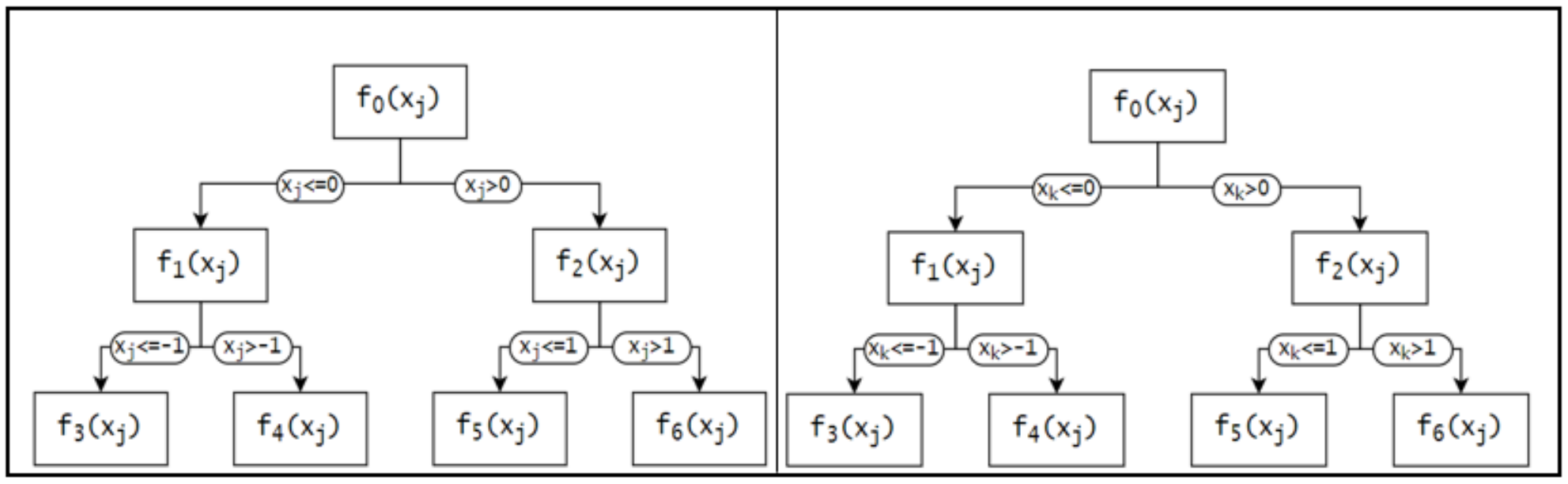

Figure 4-1. Main-effect tree $T(x_j)$ (left) and interaction-effect tree $T(x_j, x_k)$ (right)

Specifically, let $\ell(y, g(\boldsymbol{x}))$ be the loss function (squared error for continuous response and log-loss for binary response) with $L = \frac{1}{n}\sum_{i=1}^{n} \ell(y_i, g(\boldsymbol{x}_i))$, the total loss on the training set. To fit the main effect or interaction, at each iteration $m$, we want to add a new tree to the current model $g_{m-1}(\boldsymbol{x})$. Denote the new tree as $T_m(\boldsymbol{x}_s)$m where $s = j$ for main-effect or $s = (j, k)$ for interaction. Applying second-order Taylor expansion to the loss function, we get

$$L \approx \frac{1}{n}\sum_{i=1}^{n} \ell(y_i, g_{m-1}(\boldsymbol{x}_i)) + \frac{1}{n}\sum_{i=1}^{n} G_{i,m} T_m(\boldsymbol{x}_{i,s}) + \frac{1}{2n}\sum_{i=1}^{n} H_{i,m} T_m^2(\boldsymbol{x}_{i,s}),$$

where $G_{i,m} = \frac{\partial \ell(y_i, g_{m-1})}{\partial g_{m-1}}$, $H_{i,m} = \frac{\partial^2 \ell(y_i, g_{m-1})}{\partial g_{m-1}^2}$. Define the “pseudo-response” $z_{i,m} = \frac{-G_{i,m}}{H_{i,m}}$. Now, minimizing the approximate loss is equivalent to solving a weighted least square problem

$$\min_{s \in S} \sum_{i=1}^{n} H_{i,m} \left(z_{i,m} - T_m(\boldsymbol{x}_{i,s})\right)^2,$$

where $S = \{1, \dots, p\}$ for fitting main-effects or selected top interaction pairs set for fitting interactions (to be described later). Table 4-1 summarizes the algorithm for fitting main effects and interactions.

Table 4-1. Algorithm 1 for Fitting Main Effects and Interactions

| **Algorithm 1: FitMain/FitInt** | |
|---|---|
| | *Input: train, validation, initial $g_0(\boldsymbol{x})$, $S = \{1, \dots, p\}$ or selected top interaction pairs set* |
| | for $m$ =1 to $M$ do: |
| | calculate first, second order derivatives $G_{i,m}$, $H_{i,m}$ and pseudo-response $z_{i,m}$ |
| | for $s$ in $S$ do: |
| | fit a tree $T_m(\boldsymbol{x}_s)$ to $z_{i,m}$ with weights $H_{i,m}$ |
| | calculate $SSE_s = \sum_{i=1}^{n} H_{i,m-1} \left(z_{i,m} - T_m(\boldsymbol{x}_{i,s})\right)^2$ |
| | $s^* = \arg\min_{s \in S} SSE_s$ |
| | $g_m(\boldsymbol{x}) = g_{m-1}(\boldsymbol{x}) + \lambda T_m(\boldsymbol{x}_{s*})$ # $\lambda$ is learning rate |
| | calculate validation set loss at $m$-th iteration $L_m = \frac{1}{n'}\sum_{i=1}^{n'} \ell(y_i, g_m(\boldsymbol{x}_i))$, $n'$ is sample size |
| | if $L_{m-d} < \min L_{m-d+1:m}$: # stop if no improvement in last $d$ iterations |
| | $g(\boldsymbol{x}) = g_{m-d}(\boldsymbol{x})$ # the final $g(\boldsymbol{x})$ is rolled back to $d$ iteration earlier |
| | break |

### 4.1.2 New Interaction Detection Algorithm

As noted earlier, fitting all possible interactions pairs is computationally intensive. Hu et al. (2022) also proposed a new interaction filtering method using interaction-effect trees to select the top interaction pairs. Compared to the simple four-quadrant technique used in EBM, interaction-effect trees are more flexible and provide a better fit.

Table 4-2. Algorithm 2 for Interaction Filtering

| **Algorithm 2: FilterInt** | |
|---|---|
| | *Input: train,* $g(\boldsymbol{x})$ |
| | calculate first and second order derivatives $G_i$ and $H_i$ and pseudo-response $z_i$ |
| | for $j = 1$ to $p - 1$ do: |
| | for $k = j + 1$ to $p$ do: |
| | fit an interaction effect tree $T(x_j, x_k)$ to $z_i$, with weights $H_i$ |
| | calculate $SSE_{jk} = \sum_{i=1}^{n} H_i \left(z_i - T(x_{ij}, x_{ik})\right)^2$ |
| | fit an interaction effect tree $T(x_k, x_j)$ to $z_i$, with weights $H_i$ |
| | calculate $SSE_{kj} = \sum_{i=1}^{n} H_i \left(z_i - T(x_{ik}, x_{ij})\right)^2$ |
| | rank all pairs by $\min(SSE_{jk}, SSE_{kj})$ in ascending order # choose the smaller SSE |
| | select top $q$ pairs $\{(j_1, k_1), \ldots, (j_q, k_q)\}$ |
| | output $S = \{(j_1, k_1), \ldots, (j_q, k_q)\} \cup \{(k_1, j_1), \ldots, (k_q, j_q)\}$ |

One first calculates the pseudo-residuals after fitting the main effects. Then for each pair of variables $(x_j, x_k)$, two interaction trees are fitted to the pseudo-residuals, $T(x_j, x_k)$ and $T(x_k, x_j)$, by alternating their roles as modeling variable or splitting variable. By default, Hu et al. (2022) use trees with maximum depth 2 and use linear B-splines with 5 knots (including two boundary knots) to transform the modeling variable. The smaller SSE of the two fitted interaction trees is selected to measure the explainability power of the pair. Then all pairs are ranked to select the top $q$ pairs and output $2q$ (modeling variable, splitting variable) combinations as our "selected top interaction pairs set" S. See Table 4-2 for the detailed algorithm.

### 4.1.3 Full Algorithm

The full algorithm is summarized in Table 4-3. One first initializes $g(\boldsymbol{x})$ to be the overall mean for continuous response or logit of overall probability for binary response. Then the FitMain, FilterInt and FitInt functions are repeatedly called (in that order) until the early stop iterations for the main effect stage ($M_{main_stop}$) and interaction stage ($M_{int_stop}$) are both 0, or the maximum number of iteration rounds ($R$) has been reached. Note the FitMain and FitInt functions use the output model from the previous stage as initial value $g_0(\boldsymbol{x})$, to continue adding main effects/interactions to it. Therefore, it is another layer of boosting. The benefit of this iterative fitting method will be illustrated in Section 5.

Table 4-3. GAMI-Lin-T Algorithm

| **Algorithm 3: GAMI-Lin-T** | |
|---|---|
| | *Input: train, validation* |
| | Initialize $g(\boldsymbol{x}) = g_0(\boldsymbol{x})$ to be overall mean (continuous) or overall logit (binary) |
| | for $r = 1$ to $R$ do: |
| | FitMain(train, validation, $g(\boldsymbol{x})$, {1,…,p}), record early stopping rounds $M_{main_stop}$ |
| | $S$=FilterInt(train, $g(\boldsymbol{x})$) |
| | FitInt(train, validation, $g(\boldsymbol{x})$, $S$), record early stopping rounds $M_{int_stop}$ |
| | if $M_{main_stop} == 0$ and $M_{int_stop} == 0$: |
| | break # stop if both main and interaction stages stop with 0 iterations |

While the iterative fitting method and more flexible interaction filtering model are beneficial, one may be concerned about the computational cost. Hu et al. (2022) used efficient algorithms and applied high-performance computing tools to speed up the code (see Appendix A). We tested the algorithm on data sets with 100K to 10M number of observations and 50 variables. The fitting time ranges from well within 1 minute to a few minutes, and we found the computational speed to be satisfactory.

## 4.2 Purifying Main Effects and Interactions

For the purposes of uniqueness and interpretation, the main effects and interactions must be orthogonal. Even though they are estimated separately (and interactions are fit on the residuals from main effects), they may not exactly satisfy the hierarchical orthogonality constraints $\int g_{jk}(x_j, x_k) g_\ell(x_\ell) w_{jk}(x_j, x_k) dx_j dx_k = 0,\ \ \ell = j, k$ where $w_{jk}(x_j, x_k)$ is the marginal density function of $x_j$ and $x_k$ (see Hooker 2007). Therefore, Hu et al. (2022) recommend a "purification" process after the model is fit to enforce this property.

For a given $x_j$, sum all main-effect trees $T_m(x_j)$ to form the initial estimate of $g_j(x_j)$. Similarly, for a given pair $(x_j, x_k)$, sum all interaction effect trees $T_m(x_j, x_k)$ and $T_m(x_k, x_j)$ to form initial estimate of $g_{jk}(x_j, x_k)$. Obtain the pseudo-predictions on the training data as $\hat{y}_i^{jk} = g_{jk}(x_{ij}, x_{ik}), i = 1, \ldots, n$. Then fit an additive model $h_j(x_j) + h_k(x_k)$ to $\hat{y}_i^{jk}$'s using B-spline transformed $x_j$ and $x_k$ variables (other ways of fitting additive models can also be used). One gets the orthogonalized interactions as $\tilde{g}_{jk}(x_j, x_k) = g_{jk}(x_j, x_k) - h_j(x_j) - h_k(x_k)$. The subtracted main effects are added to $g_j(x_j)$ and $g_k(x_k)$: $\tilde{g}_j(x_j) = g_j(x_j) + h_j(x_j)$ and $\tilde{g}_k(x_k) = g_k(x_k) + h_k(x_k)$. With this approach, the final orthogonalized interaction and main effects are hierarchically orthogonal. A proof is provided in Appendix B. Other approaches of purification can also be used. For example, Lengerich (2020) proposed a method which iteratively removes the means projected onto a lower dimension covariate space, until all the lower dimensional means are 0. This is similar to what is done with GAMI-Lin-T, except all the lower dimension means are simultaneously estimated instead of iteratively.

This purification step is not done in EBM. GAMI-Net implements orthogonality by clarity regularization. It is a constraint which means main-effects and interactions may not be exactly orthogonal, creating some difference between GAMI-Net/EBM and GAMI-Lin-T- (see Section 6).

## 4.3 Similarities and Differences between EBM and GAMI-Lin-T

Both are tree-based algorithms, and they share several similarities: estimating main effects and interactions in separate stages, filtering top interactions, and fitting model in an additive way using simple base learners. However, there are some key differences as described below:

1. GAMI-Lin-T uses linear/spline-model trees as base learners in fitting main effects and second order interactions. These are more flexible and require fewer splits and fewer number of trees to capture a complex function. In general, they lead to less overfitting and hence they have better generalization performance. This was observed this in another study (Aramideh et al., 2022)
2. There is a new interaction filtering method using spline-model trees. Even though the simple 4-quadrant model used in FAST works well, model-based tree can capture interaction patterns better and rank the interaction effects more accurately in some cases (see Section 5).
3. GAMI-Lin-T uses an iterative fitting method to fit the main effects and interactions, instead of the two-stage fitting method used in EBM. This has two advantages listed below, and they lead to performance improvement if we iterate.
   - When main effects and interaction terms are not orthogonal, fitting main effects and interaction terms cannot be done in the naïve two-stage way. As an analogy, if we think of the main effects and interaction terms as two correlated predictors $x_1, x_2$ (but not perfectly collinear), we cannot just fit $x_1$ first and then fit $x_2$ using the residuals; instead, we need to iteratively fit one predictor at a time until convergence (or fit the two simultaneously). Otherwise, we will find bias and worse model fit. We will demonstrate this in Section 5.
   - Some weaker interaction terms may be missed in the initial round of filtering. By iterating, one can capture the missed ones in the subsequent iterations. Therefore, it is better at capturing all true interactions. We will also demonstrate this in Section 5.
4. Instead of the round-robin training method used in EBM which cycles through all variables (or interaction pairs) at each round, GAMI-Lin-T only chooses the most important variable (or interaction pair) to model at each iteration. Thus, it overfits less on the non-significant variables (interaction pairs).
5. Finally, there is the purification step, so the main effects and interactions are hierarchically orthogonal based on the functional-ANOVA concept in Hooker (2007).

# 5 Comparison of Algorithms on Simulation Data

We used several simulation cases to compare XGBoost, GAMI-Lin-T, GAMI-Net, and EBM in terms of model performance and model interpretation.

## 5.1 Simulation Setup

We considered the following four models:

1. $g(\boldsymbol{x}) = \sum_{j=1}^{5} x_j + \sum_{j=6}^{8} 0.5x_j^2 + \sum_{j=9}^{10} x_j\, I(x_j > 0) + \sum_{j=1}^{10} \sum_{k=j+1}^{10} 0.2x_j x_k$ ;
2. $g(\boldsymbol{x}) = \sum_{j=1}^{5} x_j + \sum_{j=6}^{8} 0.5x_j^2 + \sum_{j=9}^{10} x_j\, I\left(x_j > 0\right) + 0.25x_1x_2 + 0.25x_1x_3^2 + 0.25x_4^2x_5^2 +$ $\exp\left(\frac{x_4x_6}{3}\right) + x_5x_6 I(x_5 > 0)(x_6 > 0) + clip(x_7 + x_8,\ -1,\ 0) + clip(x_7x_9,\ -1,\ 1)$ $+ I(x_8 > 0)\, I(x_9 > 0)$, where $clip(x, a, b)$ are the cap and floor functions;

3. $g(\boldsymbol{x}) = \sum_{j=1}^{5} x_j + \sum_{j=6}^{8} 0.5x_j^2 + \sum_{j=9}^{10} x_j\, I\big(x_j > 0\big) + 0.25x_1^2x_2^2 + 2(x_3 - 0.5)_+(x_4 - 0.5)_+ + 0.5\sin(\pi x_5)\sin(\pi x_6) + 0.5\sin\big(\pi(x_7 + x_8)\big)$;
4. $g(\boldsymbol{x}) = \sum_{j=1}^{5} x_j + \sum_{j=6}^{8} 0.5x_j^2 + \sum_{j=9}^{10} x_j\, I\big(x_j > 0\big) + x_1x_2 + x_1x_3 + x_2x_3 + 0.5x_1x_2x_3 + x_4x_5 + x_4x_6 + x_5x_6 + 0.5\, I(x_4 > 0)x_5x_6$;

Model 1 contains a total of 45 interactions, and we wanted to see if all of them can be captured. Model 2 has eight different forms of interactions. Model 3 includes the oscillating sine functions, which would be hard to capture by the 4-quandrat approximation used in FAST. Model 4 contains two 3-way interactions. We included it here to assess the performance of GAMI models. In practice, they will capture the projection of 3-order interactions into one and two-dimensions. If the remaining third-order interaction is large, they will not perform as well as XGBoost.

For each model form, we simulated 20 predictors $x_1 \sim x_{20}$ from multivariate Gaussian distribution with mean 0, variance 1 and equal correlation $\rho$. Only the first 10 variables, $x_1 \sim x_{10}$, are important in the model, and the rest (*redundant* variables) do not affect performance although they will be relevant when $\rho > 0$. We also simulated 10 additional variables $x_{21} \sim x_{30}$ which were independent of the first 20 variables (*irrelevant* variables). They were also simulated from multivariate Gaussian distribution with mean 0, variance 1 and equal correlation $\rho$. So, there were 30 predictors in total. To avoid potential outliers in $\boldsymbol{x}$ to be too influential, we truncated all predictors to be within the interval $[-2.5, 2.5]$.

The response was simulated as $y = g(\boldsymbol{x}) + \epsilon,\ \epsilon \sim N(0, 0.5^2)$ for continuous case. For the binary case, the response was $Bernoulli\big(p(\boldsymbol{x})\big)$, where $p(\boldsymbol{x}) = \frac{e^{\beta_0 + g(x)}}{1 + e^{\beta_0 + g(x)}}$ and the intercept $\beta_0$ was carefully chosen to have balanced classes. Only the results for the continuous case are reported here. Readers are referred to Hu et al. (2022) for results for binary case. We considered two correlation levels $\rho = 0, 0.5$. For each model form and correlation level, we simulated datasets with two different sample sizes, 50K and 500K. Each dataset was divided into train, validation, and testing sets, with 50%, 25% and 25% sample sizes, respectively. We used training set and validation set to train and tune four models (XGBoost, EBM, GAMI-Lin-T, and GAMI-Net) and evaluated the predictive performance on the test set. To avoid the randomness related with data splitting, we repeat the data splitting 10 times with different random seeds and computed the average test set performance. The standard deviations of the 10 splits are shown in parenthesis.

Below are the tuning settings:

- For EBM, we tuned max_bins, max_interaction_bins and learning rate and fixed the number of interaction pairs to be 45 for Model 1 and 10 for the other models. We used random search with a total number of 12 trials.
- For GAMI-Lin-T, we fixed all hyper-parameters and used early stopping to tune the model. Specifically, we fixed learning rate at 0.2, maximum depth at 2, number of maximum rounds at 5, number of maximum iterations per main/interaction stage at 1000 and number of interactions to be 45 for Model 1 and 10 for the other models.
- For GAMI-Net, we used a subnet architecture of 5 layers, each with 40 neurons. Several hyperparameters were fixed: number of epochs at 200, learning rate at 0.0001, batch size at 1000, number of interactions at 45 for Model 1 and at 10 for the other models, and clarity penalty was set as 0.1. These settings were decided in consultation with the authors of the GAMI-Net paper.

- For XGBoost, we tuned maximum depth and learning rate using grid search, and we used early stopping for the number of boosting rounds.

## 5.2 Simulation Results

### 5.2.1 Continuous Case

The averages and standard deviations of test mse computed from the 10 data splitting seeds, are reported for all algorithms in Table 5-1 (uncorrelated) and Table 5-2 (correlated). The main conclusions are:

i) GAMI-Net and GAMI-Lin-T are generally competitive except when the FAST interaction filtering (in GAMI-Net) misses some interactions.
ii) EBM is comparable (as good or slightly worse) to GAMI-Lin-T and GAMI-Net when there is no correlation among the predictors, but has significantly worse performance when there is correlation;
iii) GAMI-Net and GAMI-Lin-T do better than XGBoost in cases 1-3 with up to two-way interactions; and
iv) Surprisingly, when the predictors are correlated, they can even outperform XGboost in the presence of third-order interactions.

Table 5-1. Test mses for continuous simulation cases: predictors are uncorrelated

| | **N** | $\rho$ | **XGBoost** | **GAMI-Lin-T** | **GAMI-Net** | **EBM** |
|---|---|---|---|---|---|---|
| Model 1 | 50K | 0 | 0.528 (0.018) | 0.307 (0.005) | 0.297 (0.028) | 0.320 (0.004) |
| Model 1 | 500K | 0 | 0.349 (0.005) | 0.261 (0.001) | 0.259 (0.002) | 0.263 (0.001) |
| Model 2 | 50K | 0 | 0.396 (0.008) | 0.269 (0.003) | 0.266 (0.003) | 0.293 (0.003) |
| Model 2 | 500K | 0 | 0.307 (0.003) | 0.258 (0.001) | 0.257 (0.002) | 0.261 (0.001) |
| Model 3 | 50K | 0 | 0.441 (0.007) | 0.278 (0.003) | 0.431 (0.048) | 0.321 (0.022) |
| Model 3 | 500K | 0 | 0.311 (0.003) | 0.260 (0.001) | 0.256 (0.002) | 0.263 (0.001) |
| Model 4 | 50K | 0 | 0.497 (0.042) | 0.603 (0.010) | 0.570 (0.009) | 0.685 (0.015) |
| Model 4 | 500K | 0 | 0.337 (0.011) | 0.559 (0.002) | 0.553 (0.003) | 0.570 (0.002) |

Table 5-2. Test mses for continuous simulation cases: predictors are correlated

| | **N** | $\rho$ | **XGBoost** | **GAMI-Lin-T** | **GAMI-Net** | **EBM** |
|---|---|---|---|---|---|---|
| Model 1 | 50K | 0.5 | 0.658 (0.022) | 0.318 (0.005) | 0.290 (0.008) | 0.891 (0.021) |
| Model 1 | 500K | 0.5 | 0.454 (0.008) | 0.270 (0.001) | 0.266 (0.006) | 0.690 (0.004) |
| Model 2 | 50K | 0.5 | 0.456 (0.011) | 0.275 (0.004) | 0.342 (0.029) | 0.413 (0.007) |
| Model 2 | 500K | 0.5 | 0.344 (0.003) | 0.260 (0.001) | 0.311 (0.003) | 0.355 (0.002) |
| Model 3 | 50K | 0.5 | 0.476 (0.007) | 0.286 (0.004) | 0.460 (0.006) | 0.505 (0.007) |
| Model 3 | 500K | 0.5 | 0.328 (0.004) | 0.271 (0.001) | 0.445 (0.001) | 0.461 (0.003) |
| Model 4 | 50K | 0.5 | 0.603 (0.022) | 0.372 (0.009) | 0.350 (0.007) | 0.953 (0.032) |
| Model 4 | 500K | 0.5 | 0.386 (0.007) | 0.337 (0.001) | 0.335 (0.004) | 0.764 (0.006) |

Below is a more detailed summary of the findings:

- Case 4 with third order interactions:
  - XGBoost significantly outperforms EBM for all sample sizes and all correlations.

- But XGBoost does not do as well as GAMI-Net and GAMI-Lin-T in the correlated cases. In this situation, part of the third-order interaction is confounded with first and second-order effects, so GAMI-Net and GAMI-Lin-T capture that part. It is, however, surprising that they have better predictive performance overall. Of course, this will not be the case if the third-order interaction is big.
  - Among the GAMI models, EBM performs much worse than the other two when $\rho$ = 0.5. GAMI-Lin-T and GAMI-Net are generally comparable, with GAMI-Net doing better than the other in the four different scenarios.
- Cases 1-3: XGBoost should do about the same as the GAMI models when there are only two-order interactions. two-way interactions. Surprisingly, the performance of XGBoost is significantly worse in the cases we studied.
- Overall, EBM does not perform as well as GAMI-Net or GAMI-Lin-T when the predictors are correlated.
- GAMI-Lin-T and GAMI-Net have similar performances, except for a few cases below.
  - For Model 1, 50K, $\rho$ = 0.5, GAMI-Net has 9% smaller MSE. This is likely due to neural networks being better at capturing such linear interaction effects with smaller sample sizes. As sample size increases to 500K, this difference becomes small.
  - For Model 2, $\rho$ = 0.5 and Model 3, GAMI-Lin-T is better than GAMI-Net. As we will see later, this is because the FAST interaction filtering method (used in both EBM and GAMI-Net) misses some true interactions terms.
- Comparisons of training and test data performances (not shown here) reveal that GAMI-Net has smaller train/test mse gap than other algorithms. The likely reason is that NNs are smooth and overfit less. GAMI-Lin-T overfits less than EBM and XGBoost.
- We also compared GAMI-Lin-T with a variation that trains for only one round of main-effect and interaction stage (not shown here). Results indicated that the iteration between main-effect stage and interaction stage is important, especially for correlated cases. More details will be given later.

We now turn to an examination of the identified effects in the GAMI algorithms in more detail, starting with main effects. There are 10 true main effect variables in the model, $x_1 - x_{10}$. All algorithms captured these 10 effects as the 10 most important. For the redundant or irrelevant variables, GAMI-Lin-T and GAMI-Net do the best job in assigning low importance scores to those variables. There are two reasons:

1. In the round-robin training method used in EBM, all variables are used regardless of whether they are truly important or not. However, GAMI-Lin-T selects only the best variables to model in each iteration, and it stops if model performance stops improving. In GAMI-Net, a pruning step is used, which keeps only the top $k$ −most important terms. Therefore, most non-model variables have exactly zero importance.
2. When the variables are correlated, the main-effect stage is more prone to assign importance to correlated, non-model variables. However, iterative training in GAMI-Lin-T can reverse the identification of false main effects from the first round, leading to close-to-zero importance for such redundant variables. GAMI-Net has a fine-tune stage where all main-effects and interactions are retrained simultaneously. This has the same effect as iterative training employed in GAMI-Lin-T.

In general, for $\rho = 0$, the main effect estimates for active variables ($x_1 - x_{10}$) are similar for all GAMI algorithms. EBM is wigglier due to its piecewise constant nature. GAMI-Net is smoother than GAMI-Lin-T as it is continuous at the boundaries.

For $\rho = 0.5$, Figure 5-1 shows the main effects plots for $x_1 - x_{10}$ for Model 4. Due to the correlation, part of interaction effects is captured as main effect, creating the quadratic pattern for $x_1 - x_5$. GAMI-Net and GAMI-Lin-T are very close, whereas EBM shows somewhat different patterns for $x_7 - x_{10}$. In particular, $x_9$ and $x_{10}$ are purely additive in the model and the true main effect is the hinge function $x_j\, I(x_j > 0)\, , j = 9, 10$. EBM shows a slight uptick pattern in the negative region, whereas GAMI-Lin-T and GAMI-Net are close to the true function which is flat in the negative region. This is also observed for the 500K sample size case. In general, for $\rho = 0.5$, the iterative training in GAMI-Lin-T and the fine-tuning step in GAMI-Net lead to more accurate results.

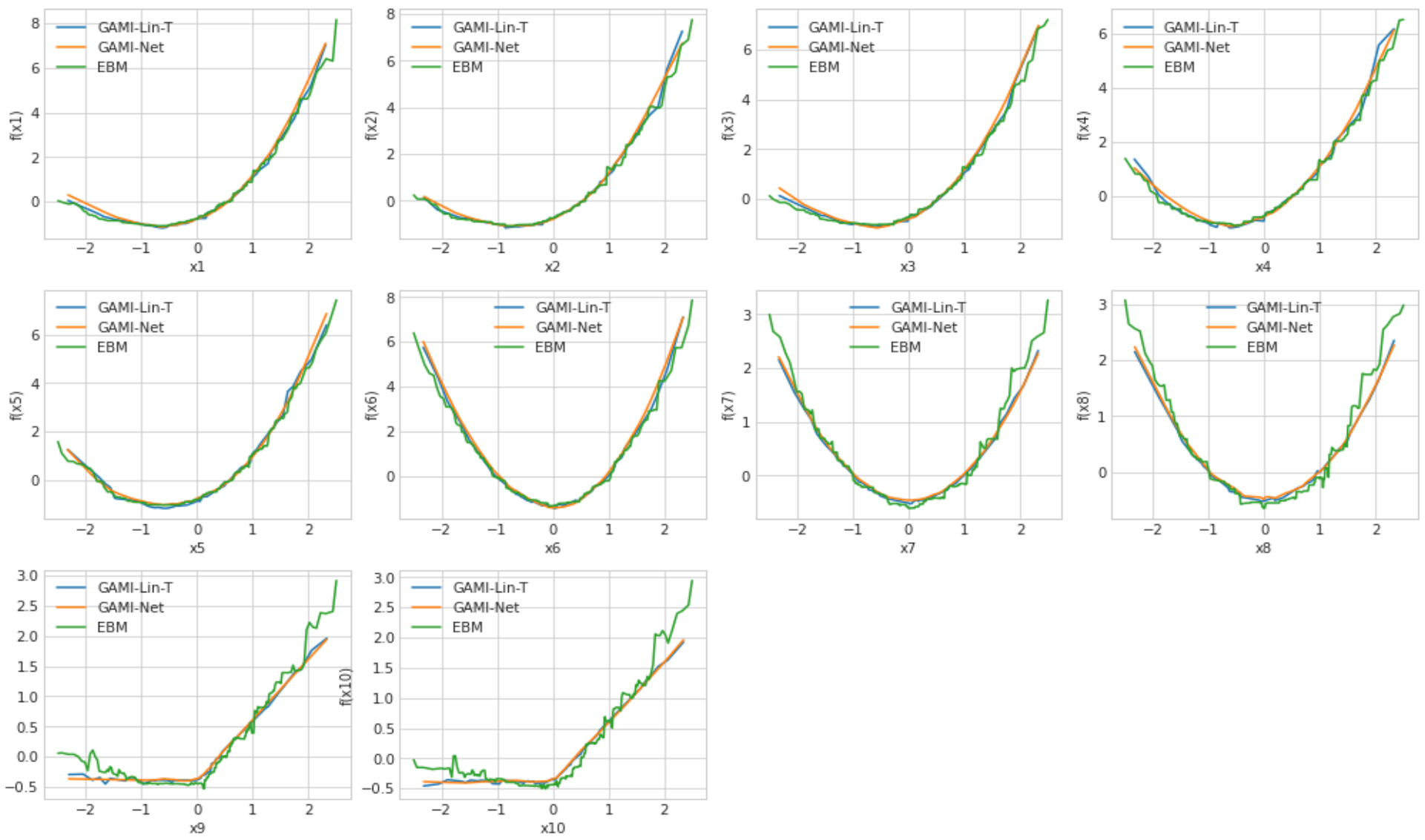

Figure 5-1. Main effect for true model variables, Model 4, ρ = 0.5, 50K

We now examine two-way interactions. The main conclusion is that the difference among the algorithms, in terms of the top interactions identified, is due to the interaction filtering methods used. In a few cases, the FAST algorithm used in EBM and GAMI-Net miss a few active interactions while the interaction tree method in GAMI-Lin-T do not miss any. More detailed results are provided below.

- For Models 1 and 4, all algorithms capture all active interaction pairs as the top ones.
- For Model 2, $\rho = 0$, all eight active interaction pairs are captured as the top.
- For Model 2, $\rho = 0.5$, EBM and GAMI-Net miss two active interaction pairs in their top 10 list: $0.25 x_1 x_2$ and $clip(x_7 + x_8, -1, 0)$, for both 50K and 500K sample sizes. An example for the 50K sample case is shown in Figure 5-2. This is due to the correlation among variables. It causes the "pure interaction" effect for these two smaller interactions to be weaker and harder to identify. Some "surrogate interactions", that are not in the true model but mimicking the true interaction pairs due to variable correlation, for the six strong interactions rank higher during interaction filtering. For GAMI-Lin-T, its second-round redoes the interaction filtering. Because interactions for the first six pairs have already been accounted for in the first round, the surrogate interaction pairs are no longer significant, and it is easy to identify the missed interactions. As a result, the second round accurately picked up the two missed interaction as the two most important interaction pairs, and GAMI-Lin-T was able to capture all eight true interaction pairs and list them as top eight correctly.

- For Model 3, $\rho = 0.5$, EBM and GAMI-Net both miss the two sine-function related interactions, $x_5$-$x_6$ and $x_7$-$x_8$, whereas GAMI-Lin-T captures all four true interactions. An example for the 50K case is shown in Figure 5-3. This is because the 4-quandrant model used in FAST algorithm cannot capture the highly nonlinear sine function well, so it misses out on these two interactions.
- For Model 3, 50K, $\rho = 0$, GAMI-Net misses two sine-function interactions (not shown here) due to the limitation of FAST algorithm mentioned earlier, explaining its worse model performance.

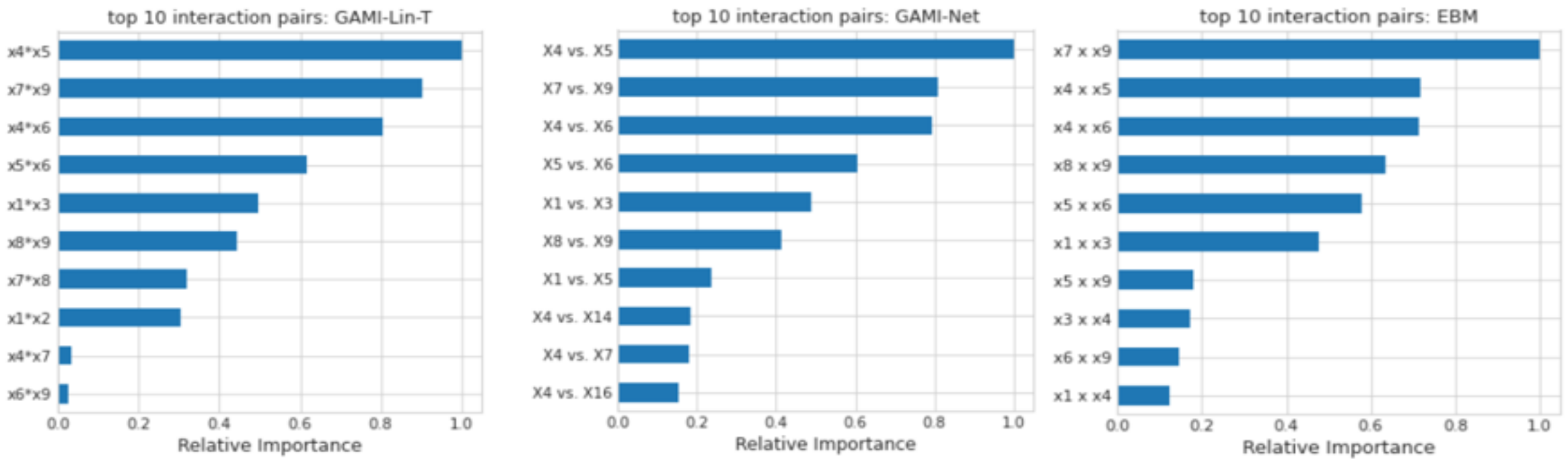

Figure 5-2. Interaction importance for Model 2, 50K and $\rho$ = 0.5

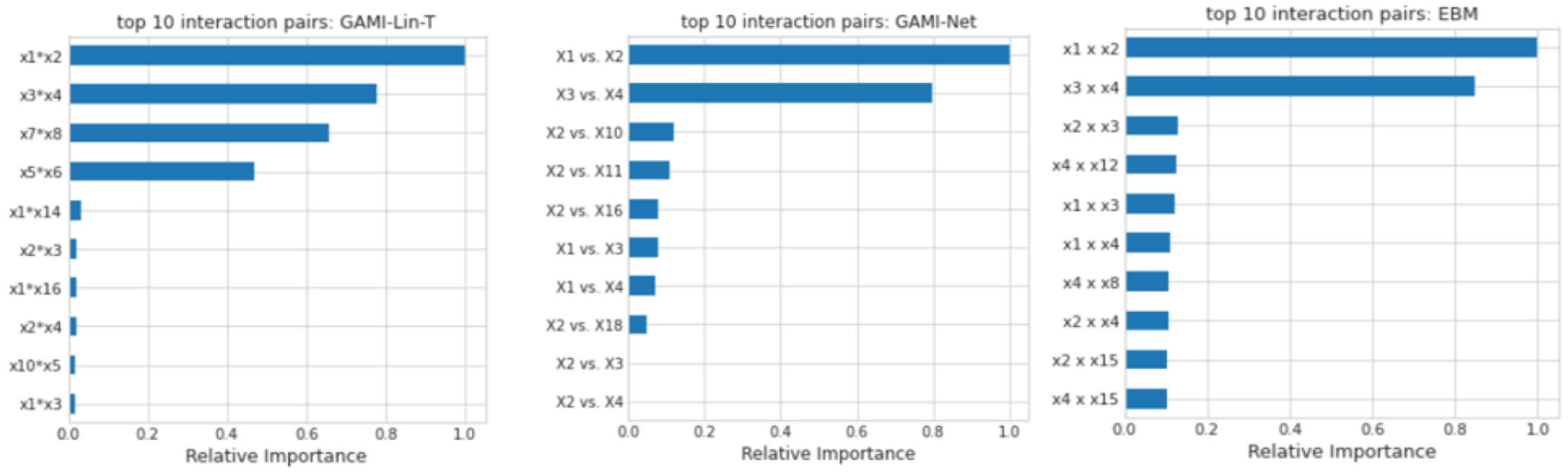

Figure 5-3. Interaction importance for Model 3, 50K and $\rho$ = 0.5

The shapes of the two-way estimated interaction effects are similar for all methods. Figure 5-4 shows the two-way interactions from GAMI-Lin-T (top), GAMI-Net (second), EBM (third) and the true model after orthogonalization (bottom). We plot $f(x_j, x_k)$ when fixing $x_k$ at a set of quantile values. Since EBM and GAMI-Net only capture six true interactions, only those six are shown. We can see they have similar patterns in general, with GAMI-Net being smoother and others showing some wiggly patterns.

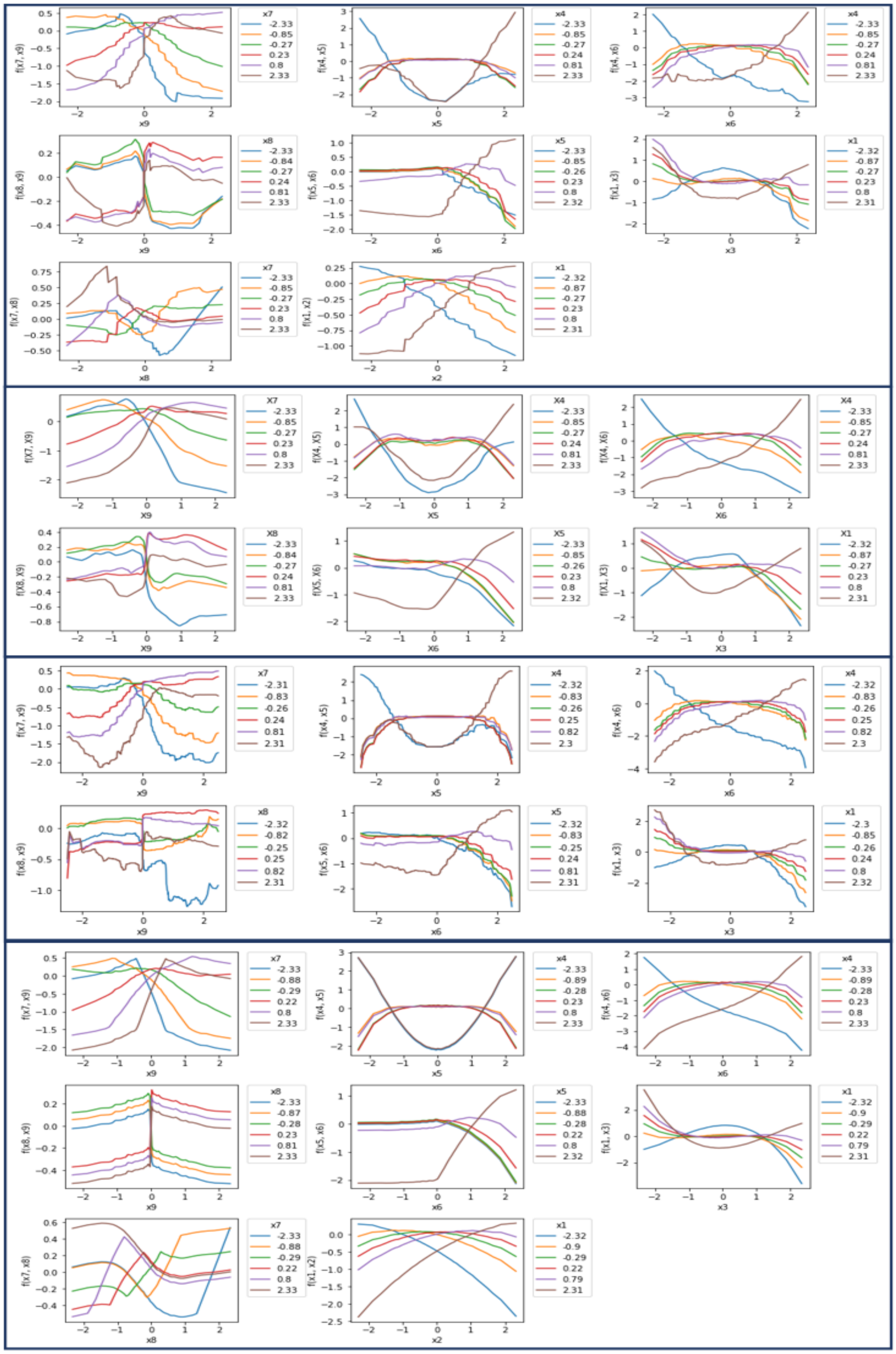

Figure 5-4. Interaction effect plot for GAMI-Lin-Tree (top), GAMI-Net (second), EBM (third) and true model (bottom), for Model 2, 50K and $\rho$ = 0.5 As GAMI-Net and EBM detect only six interactions, only those are shown.

### 5.2.2 Binary Case

The results for the binary case can be found in Hu et al. (2022). The results were qualitatively similar as the continuous case, but noisier and less significant. For example, the AUC performances among all models were very close. The main-effects and interactions captured are noisier due to smaller signal to noise ratio in binary response.

## 6 Illustration on Home Mortgage Data

This applications deals with residential mortgage accounts. The response variable is an indicator of “troubled” loan: 1 if the loan is in trouble state and 0 otherwise. The term trouble is defined as any of the following events: bankruptcy, short sale, 180 or more days of delinquency in payments, etc. The goal is to predict if a loan will be in trouble at a future prediction time based on account information at the current time (called snapshot time) and (forecasted) macro-economic information at the prediction time. The time interval between prediction time and current time is called prediction horizon (see Figure 6-1).

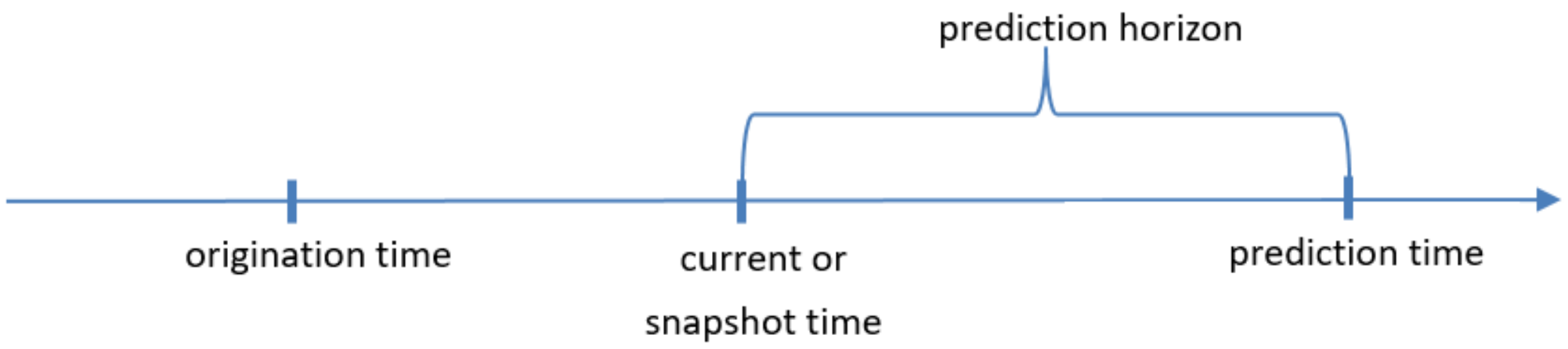

Figure 6-1. Loan origination, current (snapshot) and prediction times

There are over 50 predictors, including macro-economic variables (unemployment rate, house price index, and so on), static loan characteristic variables at the origination time (fixed 15/30 year loan, arm loan, balloon loan, etc), and dynamic loan characteristic variales (snapshot fico, snapshot delinquency status, forecasted loan-to-value ratio, etc). See Appendix C for the list of some key variables used.

We selected a subset of 1 million observations from the original dataset for one of the portfolio segments. The data was split into 50% training, 25% validation and 25% testing. Again, we split the data 10 times with different random seeds, and fitted all four algorithms: xgboost, GAMI-Net, GAMI-Lin-T, and EBM. We used the same tuning/training settings as in Section 5. The averages (and standard deviations) of testing set performance for all models are listed in Table 6-1. The performances of all the models are comparable.

Table 6-1. Test AUC as well as logloss for home lending data

| | **xgboost** | **GAMI-Lin-T** | **GAMI-Net** | **EBM** |
|---|---|---|---|---|
| **test_AUC** | 0.861 (0.003) | 0.856 (0.003) | 0.850 (0.003) | 0.853 (0.003) |
| **test_logloss** | 0.0452 (0.0005) | 0.0455 (0.0005) | 0.0460 (0.0005) | 0.0459 (0.0005) |

Figure 6-2 shows the importance ranking for the top 10 main effects. The rankings among all models are close with only small differences. For example, GAMI-Lin-T ranks horizon as $7^{th}$ important main effect, whereas GAMI-Net ranks it as $10^{th}$ and EBM does not rank it as one of the top 10 main effects; on the other

hand, EBM/GAMI-Net ranks interest only indicator as the 9th important main effect, whereas GAMI-Lin-T does not rank it as one of the top 10 main effects. The results for GAMI-Net and EBM are very close except the 10th variable is different and some slight change in the ranking. The rankings are consistent with the main-effect plots in Figure 6-3. We can see GAMI-Lin-T shows stronger main effects for a few variables compared with GAMI-Net/EBM, including snapshot fico, forecasted ltv, unemployment rate and horizon. This is consistent with the fact that horizon and unemployment rate have higher ranking in GAMI-Lin-T.

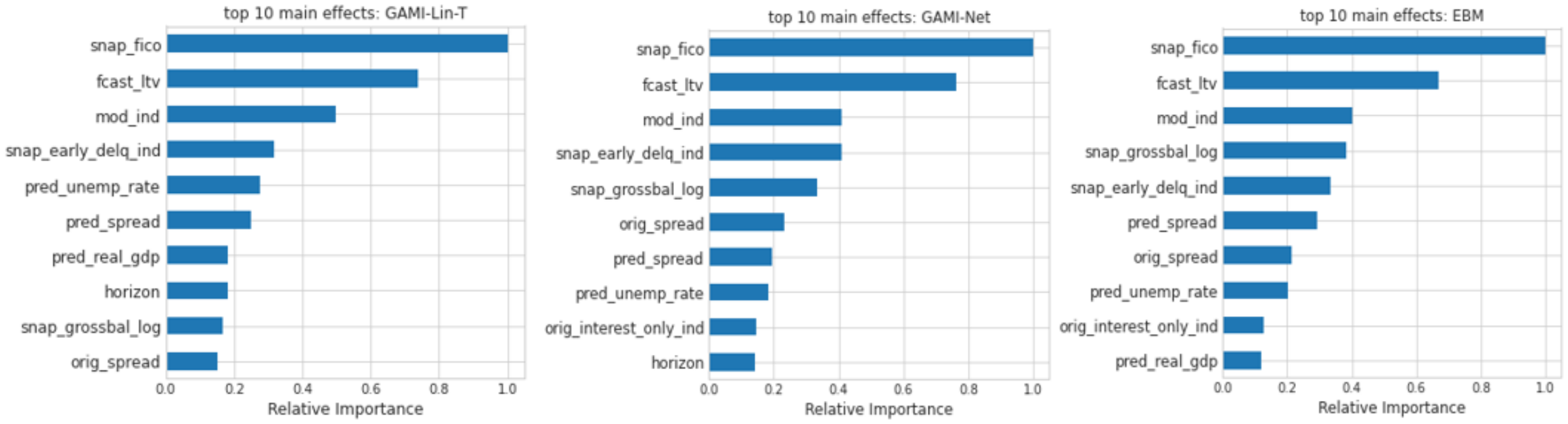

Figure 6-2. Main-effect importance for home mortgage data

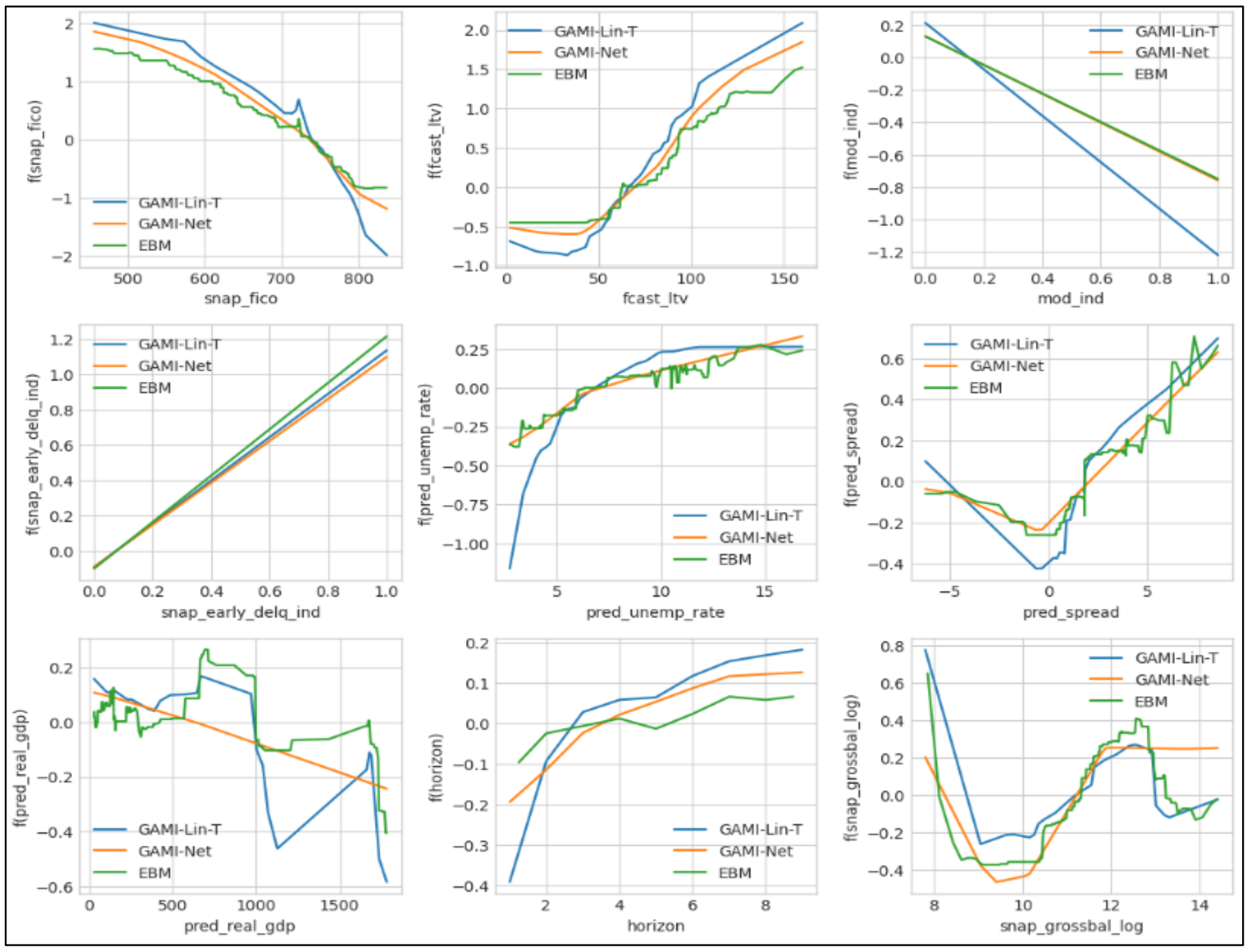

Figure 6-3. Main effect plots for home mortgage data

The difference we see in main effects among the three GAMI models are related to the different approaches to decompose main-effects and interactions. GAMI-Lin-T uses purification to gaurantee main-effects and interactions are hierarchically orthogonal, whereas EBM and GAMI-Net do not. In continuouse response cases, we found the differences to be small. However, for binary response case, the differences can be larger, causing slightly different estimates of main effects and interactions.

The top 10 interactions from GAMI-Lin-T, GAMI-Net and EBM are shown in Figure 6-4. The top four interactions from GAMI-Lin-T are: mod_ind and fico, ltv and fico, unemployment rate and fico, horizon and early delinquency indicator. Those interactions make sense from the subject-matter perspective and have been seen in other studies. The FAST algorithm in EBM did not identify the unemployment rate vs fico interaction and mod_ind vs fico interaction. On the other hand, EBM captures multiple pairs of interactions related with late delinquency indicator: most importantly, the interaction among horizon and late delinquency indicator. While this variable pair indeed has interaction, late delinquency is a very rare event (only 0.2% observations in total), and GAMI-Lin-T does not rank it as top 10. For GAMI-Net, the top 10 interactions filtered by FAST algorithm also has a lot of late delinquency indicator related interactions, but the fine tune step pruned 4 of them, keeping a total of 6 interactions. The top 2 interactions are ltv and fico, horizon and early delinquency indicator, which are high ranking interactions in all algorithms; however, it does not have unemployment rate and fico, or spread and fico interactions. Increasing the number of interactions in filtering step allows it to capture those interaction pairs, and have better model performance.

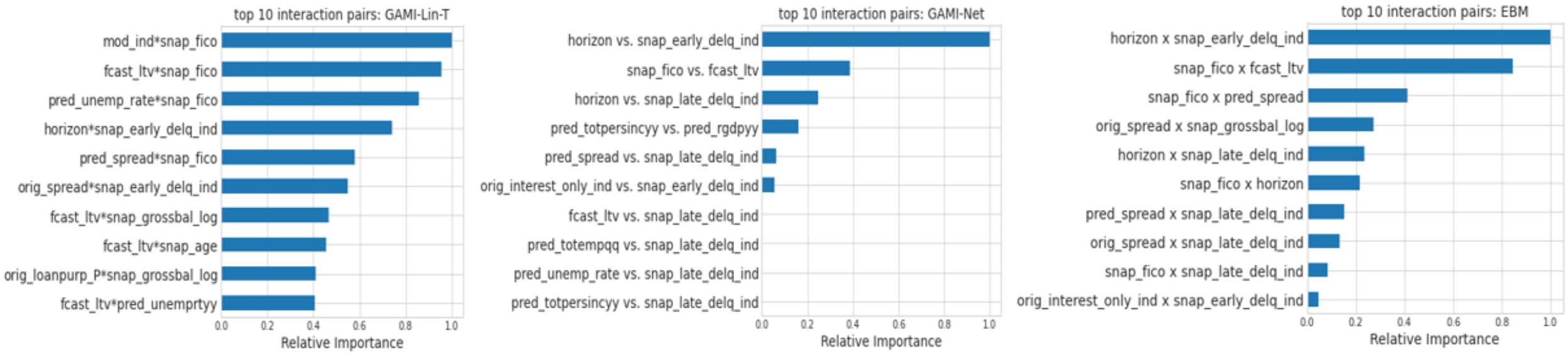

Figure 6-4. Interaction importance for home mortgage data

Figure 6-5 shows the top three of the common interaction pairs from GAMI-Lin-T and EBM, two of which are also top two in GAMI-Net. The patterns look very similar with some differences. For example, for the interaction between horizon and early delinquency indicator, GAMI-Lin-T shows that the effect of horizon is almost flat when the loan is not in early delinquency state, whereas EBM and GAMI-Net show an increasing trend. Recall that in Figure 6-3, the main-effect of horizon is flatter for EBM and GAMI-Net compared to GAMI-Lin-T. We can now see that the difference is due to how the main-effect and interactions are decomposed in each model. Particularly, the interaction effect from EBM and GAMI-Net still has some main effect on the horizon variable, this results in flatter trend for horizon in the main-effect plot. GAMI-Lin-T uses a post-hoc orthogonalization step to make sure interactions do not contain any main-effects, whereas EBM and GAMI-Net (which uses a clarity penalty) do not guarantee this.

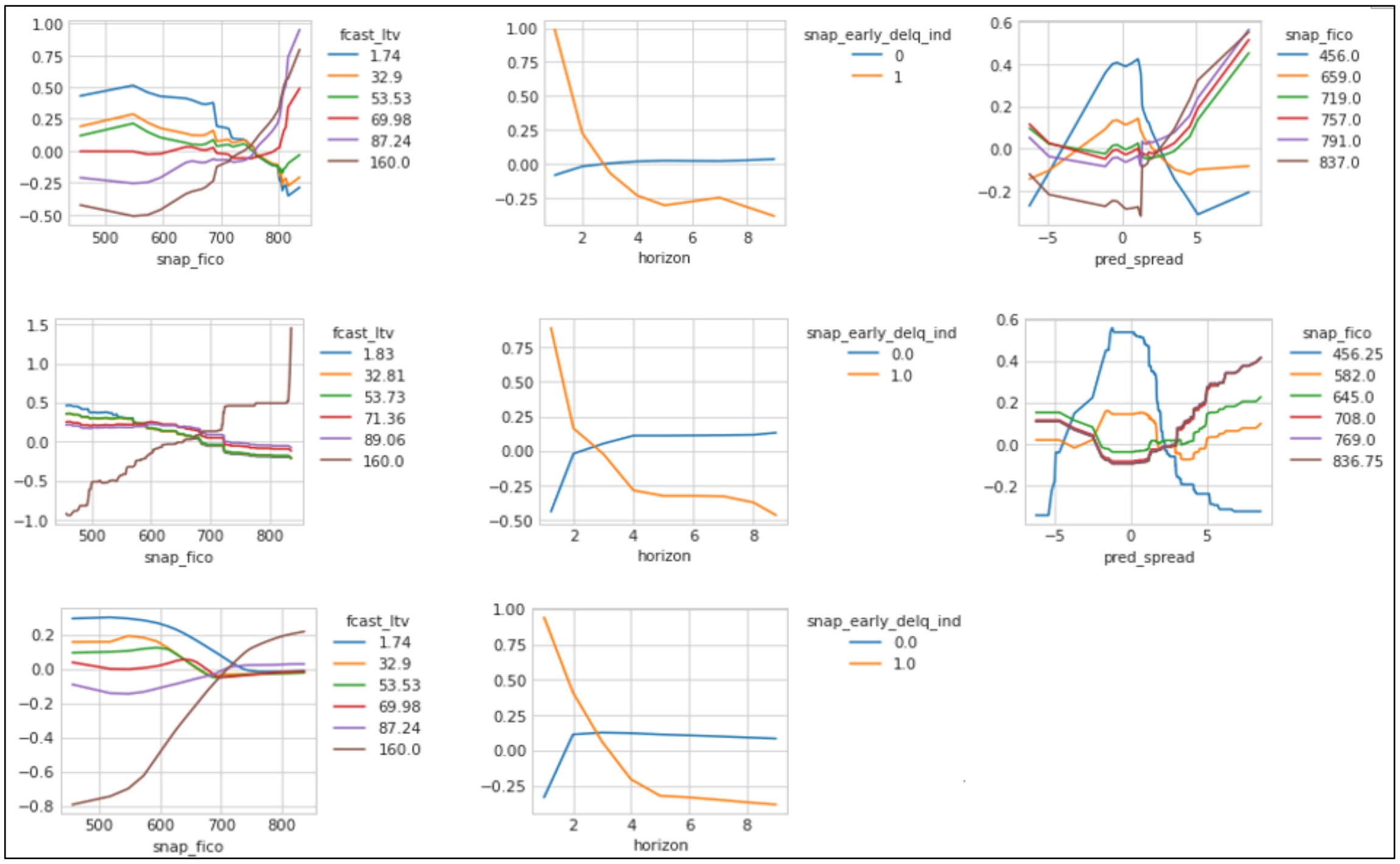

Figure 6-5. Interaction plot for home lending data, GAMI-Lin-T (top), EBM (middle), GAMI-Net (bottom)

## 7 Summary

Interpretability is an important requirement as it allows stake holders to understand and manage model risks. This paper provides an overview of the fANOVA framework and reviews two exisiting GAMI models: EBM and GAMI-Net. It also proposes a new algorithm called GAMI-Lin-T and a new interaction filtering technique. In our studies, GAMI-Lin-T performed comparably or better than EBM in terms of predictive performance and is able to identify the interactions more accurately. GAMI-Lin-T and GAMI-Net are competitive in terms of predictive performance. GAMI-Net leads to smoother estimates of the functional effects since it is continuous at the boundaries of the partitions.

## Appendix A: Computational Efficiency

Constructing model-based tree is known to be computationally expensive, because many linear models need to be fitted and evaluated in order to determine the best tree split. What is worse, GAMI-Lin-T requires fitting hundreds or even thousands of model-based trees in the boosting process. To address the computation obstacle, we have made an efficient implementation which reduces the computation by reusing intermediate results and utilizes high performance computational tools like multi-processing and Cython to speed it up.

First, to fit each model-based tree (either the main-effect tree or interaction-effect tree), we use the efficient algorithm proposed in Hu et al. (2020). Briefly, we bin the splitting variable and calculate the gram matrices, $\boldsymbol{X}^T\boldsymbol{X}, \boldsymbol{X}^T\boldsymbol{z}$, for each bin as intermediate results. Then in each tree node, we only need to find all the bins which fall into that node and sum over the corresponding binned gram matrices to obtain the gram matrix, instead of computing it from scratch. This reduces the computation cost tremendously when sample size $n$ is large since the most computation cost is in calculating the gram matrices in our applications ($n \gg p$). Moreover, only the pseudo-response $\boldsymbol{z}$ changes while the predictors stay fixed from iteration to iteration, so we can reuse the gram matrices for $\boldsymbol{X}^T\boldsymbol{X}$ and only updating the gram matrices for $\boldsymbol{X}^T\boldsymbol{z}$**.** This is fast because $\boldsymbol{z}$ is one-dimensional.

In addition, we use high performance computational tools to speed it up. The gram calculation, loss evaluation function, prediction function and solver for the ridge regression are all written in Numba or Cython, which is compiled into C code and has the speed of C. These functions are further parallelized by joblib and openmp. So, the final algorithm we have is highly optimized and parallelized.

Table A-1 shows the timing for fitting a GAMI-Lin-T model to a simulated binary response data with $n =$100K/1M/10M observations and $p = 50$ variables. The data is divided into 70% training and 30% validation, and a GAMI-Lin-T model with a particular hyper-parameter configuration (max_depth=2, ntrees=100, npairs=10, nknots=6, nrounds=1) is fitted to obtain the timing. Since the timing of GAMI-Lin-T model varies depending on how many rounds and number of trees are fitted, it is useful to show the time for each tree iteration. Table A-1 shows the average time per tree in main-effect stage and interaction stage, time for interaction filtering and total fitting and prediction time. For small data with 100K observations, it is very fast, takes less than 0.1 seconds to fit one tree. For medium data with 1M rows, it takes 0.1-0.2 seconds to fit one tree. For large data with 10M rows, it takes less than 0.7 seconds to fit one tree for nthreads=20 and less than 1.2 seconds for nthreads=10. Regarding interaction filtering, it takes only 2 second to filter all 2500 pairs of variables for the 100K data, 6-9 seconds for the 1M data and 52-75 seconds for the entire 10M data. A lot of times, we find using just a 1M subsample to filter interactions is sufficient (since the interaction model is only a two-variable model), but even with the entire 10M data, the filter speed is still acceptable. In terms of total fitting time, for the largest 10M data, a typical GAMI-Lin-T with a few hundred trees for both main-effect stage and interaction stage can be done around 10 minutes. The prediction speed is even faster, taking less than 10 seconds for the 10M data.

Table A-1. Computational Times for GAMI-Lin-T

| **n** | **p** | **nthreads** | **main stage (s/tree)** | **int-filter (s)** | **int-stage (s/tree)** | **total fit (s)** | **predict (s)** |
|---|---|---|---|---|---|---|---|
| 100K | 50 | 10 | 0.08 | 2.2 | 0.06 | 18 | 0.15 |
| 100K | 50 | 20 | 0.08 | 2.0 | 0.06 | 18 | 0.22 |
| 1M | 50 | 10 | 0.18 | 9.0 | 0.12 | 44 | 1 |
| 1M | 50 | 20 | 0.15 | 6.0 | 0.1 | 36 | 1.1 |
| 10M | 50 | 10 | 1.20 | 75 | 0.82 | 312 | 9.5 |
| 10M | 50 | 20 | 0.70 | 52 | 0.53 | 224 | 6.5 |

## Appendix B: Proof of Hierarchical Orthogonality

**Proposition**: Given a function $f(\boldsymbol{x})$, $\boldsymbol{x} = (x_1, \dots x_p)$, if $f_{-1}(\boldsymbol{x}_{-1}), \dots, f_{-p}(\boldsymbol{x}_{-p})$ are solutions that minimize $\int \left(f(\boldsymbol{x}) - \sum_{j=1}^{p} f_{-j}(\boldsymbol{x}_{-j})\right)^2 w(\boldsymbol{x}) d\boldsymbol{x}$, then the residual $\tilde{f}(\boldsymbol{x}) = f(\boldsymbol{x}) - \sum_{j=1}^{p} f_{-j}(\boldsymbol{x}_{-j})$ is orthogonal with any lower dimensional function $h(\boldsymbol{x}_{-j})$, ie, $\int \tilde{f}(\boldsymbol{x}) h(\boldsymbol{x}_{-j}) w(\boldsymbol{x}) d\boldsymbol{x} = 0$ for any $h(\boldsymbol{x}_{-j})$ and any $1 \leq j \leq p$.

**Proof**: Suppose there exists some $h(\boldsymbol{x}_{-j})$ such that $b = \int \tilde{f}(\boldsymbol{x}) h(\boldsymbol{x}_{-j}) w(\boldsymbol{x}) d\boldsymbol{x} > 0$, then consider the new solution to the least weighted sum of squared problem, $\int \left(f(\boldsymbol{x}) - \left(\sum_{j=1}^{p} f_{-j}(\boldsymbol{x}_{-j}) + \lambda h(\boldsymbol{x}_{-j})\right)\right)^2 w(\boldsymbol{x}) d\boldsymbol{x} = \int \left(\tilde{f}(\boldsymbol{x}) - \lambda h(\boldsymbol{x}_{-j})\right)^2 w(\boldsymbol{x}) d\boldsymbol{x} = \int \left(\tilde{f}(\boldsymbol{x})\right)^2 w(\boldsymbol{x}) d\boldsymbol{x} + \lambda^2 \int h^2(\boldsymbol{x}_{-j}) w(\boldsymbol{x}) d\boldsymbol{x} - 2\lambda \int \tilde{f}(\boldsymbol{x}) h(\boldsymbol{x}_{-j}) w(\boldsymbol{x}) d\boldsymbol{x} = \int \left(\tilde{f}(\boldsymbol{x})\right)^2 w(\boldsymbol{x}) d\boldsymbol{x} + \lambda^2 \int h^2(\boldsymbol{x}_{-j}) w(\boldsymbol{x}) d\boldsymbol{x} - 2\lambda b$. We can choose any small positive $\lambda < \frac{2b}{\int h^2(x_{-j}) w(x) dx}$, such that $\int \left(\tilde{f}(\boldsymbol{x}) - \lambda h(\boldsymbol{x}_{-j})\right)^2 w(\boldsymbol{x}) d\boldsymbol{x} < \int \left(\tilde{f}(\boldsymbol{x})\right)^2 w(\boldsymbol{x}) d\boldsymbol{x}$, this leads to contradiction. Similarly, we can prove the case when $\int \tilde{f}(\boldsymbol{x}) h(\boldsymbol{x}_{-j}) w(\boldsymbol{x}) d\boldsymbol{x} < 0$.

## Appendix C: Variables Used in the Home Mortgage Application

Table C-1.Variables for home mortgage data

| Variable | Definition |
|---|---|
| horizon | prediction horizon (difference between prediction time and snapshot time) in quarters |
| snap_fico | credit score (FICO score) at snapshot time |
| orig_fico | credit score (FICO score) at loan origination |
| snap_ltv | loan to value (ltv) ratio at snapshot time |
| fcast_ltv | loan to value (ltv) ratio forecasted at prediction time |
| orig_ltv | loan to value ratio at origination |
| orig_cltv | combined ltv at origination |
| snap_early_delq_ind | early delinquency (no min payments for a few months) indicator: 1 means loan has early delinquency status at snapshot time; 0 means loan is current or has late delinquency status. 7.7% observations are early delinquent. |
| snap_late_delq_ind | late delinquency indicator (loan is delinquent for longer time, close to default) indicator: 1 means loan has late delinquency status at snapshot time; 0 means loan is current or has early delinquency status. Only 0.2% observations are late delinquent. |
| pred_loan_age | age of loan (in months) at prediction time |
| snap_gross_bal | gross loan balance at snapshot time |
| orig_loan_amt | total loan amount at origination time |
| pred_spread | spread (difference between note rate and market mortgage rate) at prediction time |
| orig_spread | spread at origination time |
| orig_arm_ind | Indicator: 1 if loan is adjustable-rate mortgage (ARM); 0 otherwise |
| pred_mod_ind | modification indicator: 1 means prediction time before 2007Q2 (financial crisis); 0 if after |
| pred_unemp_rate | unemployment rate at prediction time |
| pred_hpi | house price index (hpi) at prediction time |
| orig_hpi | hpi at origination time |
| pred_home_sales | home sales data at prediction time |
| pred_rgdp | real GDP at prediction time |
| pred_totpersinc_yy | total personal income growth (from year before prediction to prediction time) |